\pdfoutput=1

\documentclass[11pt]{article}

\usepackage[review]{acl}
\usepackage{graphicx}

\usepackage{times}
\usepackage{latexsym}

\usepackage[T1]{fontenc}

\usepackage[utf8]{inputenc}

\usepackage{microtype}

\title{A Novel DeBERTa-based Model for Financial Question Answering Task}
\author{Yanbo J. Wang, Yuming Li, Hui Qin, Yuhang Guan and  Sheng Chen \\
        LYZD-FinTech Co., LTD \\ Chaoyang, Beijing, China}

\begin{document}
\maketitle
\begin{abstract}
As a rising star in the field of natural language processing, question answering systems (Q\&A Systems) are widely used in all walks of life. Compared with other scenarios, the application in financial scenario has strong requirements in the traceability and interpretability of the Q\&A systems. In addition, since the demand for artificial intelligence technology has gradually shifted from the initial computational intelligence to cognitive intelligence, this research mainly focuses on the financial numerical reasoning dataset - FinQA. In the shared task, the objective is to generate the reasoning program and the final answer according to the given financial report containing text and tables. We use the method based on DeBERTa pre-trained language model, with additional optimization methods including multi-model fusion, training set combination on this basis. We finally obtain an execution accuracy of 68.99 and a program accuracy of 64.53, ranking No. 4 in the 2022 FinQA Challenge.
\end{abstract}

\section{Introduction}
With the development of Internet technology and the popularization of informatization, the Internet industry has developed exponentially. At the same time, with the steady improvement of computer hardware and software, the demand for artificial intelligence-related work has gradually shifted from initial computational intelligence to cognitive intelligence.

In natural language processing (NLP), question answering system (Q\&A Systems) has become a popular research topic recently, and its application fields include medical \cite{jiang2021research}, finance \cite{zhu2021tat}, education \cite{goel2018jill}, public service \cite{sheng2020dsqa} etc.. Existing Q\&A systems are mainly divided into two mainstream categories, retrieval based Q\&A systems and generation based Q\&A systems. Among them, the retrieval based Q\&A systems refers to the retrieval of matching answers by certain rules and strategies according to the question text input by the user. The widely used methods are mainly based on semantic matching and knowledge graph. The essence of the question answering system based on knowledge graph is to extract unstructured data into usable knowledge, and express the relationship in the form of semantic web. Due to its strong reasoning and inference capability, it is highly interpretable and has been widely used in many practical scenarios \cite{zhang2018variational,yasunaga2021qa}. The generation based Q\&A systems is mainly based on the deep learning model, which can integrate the answer fragments distributed in different phrases. Through seq2seq learning, it can deal with the flexibility changes of the question more effectively than the former \cite{zhang2018nlp}.

In this paper, we propose a DeBERTa v3 large-based \cite{he2020deberta} model for financial question and answer task in FinQA dataset \cite{chen2021finqa}. We finally achieve the execution accuracy of 68.99 after fine-turning and optimization, and win the fourth prize in the FinQA challenge hosted at the Workshop on Structured and Unstructured Knowledge Integration (SUKI), NAACL 2022\footnote{https://finqasite.github.io/challenge.html}.

\section{Literature review}
Question answering system is a retrieval system that can find concise and clear answers from a large amount of data by understanding the user's natural language questions. The earliest question answering system can be traced back to the ELIZA \cite{weizenbaum1966eliza} proposed by MIT in 1966. However, this Q\&A mode is not to understand human intentions and thoughts, but to generate new combinations of words and sentences by analyzing the typed words and combinations of specific words and sentences, which does not have a complete knowledge base and logical structure. With the rapid development of computer intelligence, automatic question answering systems have been gradually applied in all walks of life \cite{chlebowski2022abductive,krishna2021hurdles,noraset2021wabiqa,alzubi2021cobert}. The mainstream system technologies can be roughly divided into rule-based methods and semantic analysis-based methods.

The rule-based question answering method matches the question text by establishing a template library in advance, and converts the question text into the corresponding data query statement through the preset query template. True knowledge \cite{tunstall2010true} took the lead in proposing to realize the logical search of different problems through a large number of manually formulated templates. On the basis of the previous foundation, Abujabal et al. \cite{abujabal2017quint} proposed QUINT (an interpretable question answering over knowledge bases.), which uses the existing corpus data to train an automatic learning template on this basis, and converts the question into a query language through the template. Subsequently, Cocco et al. \cite{cocco2019machine} proposed an object-oriented question answering system to learn SPARQL templates with the help of data sets in the form of RDF (Resource Description Framework).

The question answering method based on semantic parsing first converts the natural language into a semantic representation, and then generates the corresponding data query, and returns the queried answer through knowledge base. Semantic parsing-based question answering methods can be divided into three categories according to different implementation principles: grammar-based semantic parsing \cite{krishna2021hurdles}, graph-based semantic parsing \cite{sun2020sparqa}, and neural network-based semantic parsing \cite{chen2019bidirectional}.

\section{Task description}
The task can be summed up as, to provide a financial report or financial document containing the textual description  $E$ and the structured table $T$. The objective is to generate a reasoning program $G$, in which $G$ contains several program tokens defined by Domain Specific Language (DSL), and finally get the answer $A$ based on the given question $Q$.
    $$P(A|T, E, Q) = \sum P(G_{i}|T, E, Q)$$
In which the DSL consists of 6 math operations (\textit{\textbf{add, subtract, multiply, divide, greater, exp}}) and 4 tabular operations (\textit{\textbf{table-max, table-min, table-sum, table-average}}), each of which contains an argument list $args_n$. The definitions of all operations are shown in Figure 1:

\begin{figure}[htbp]
    \centering
    \includegraphics[width=7.8cm]{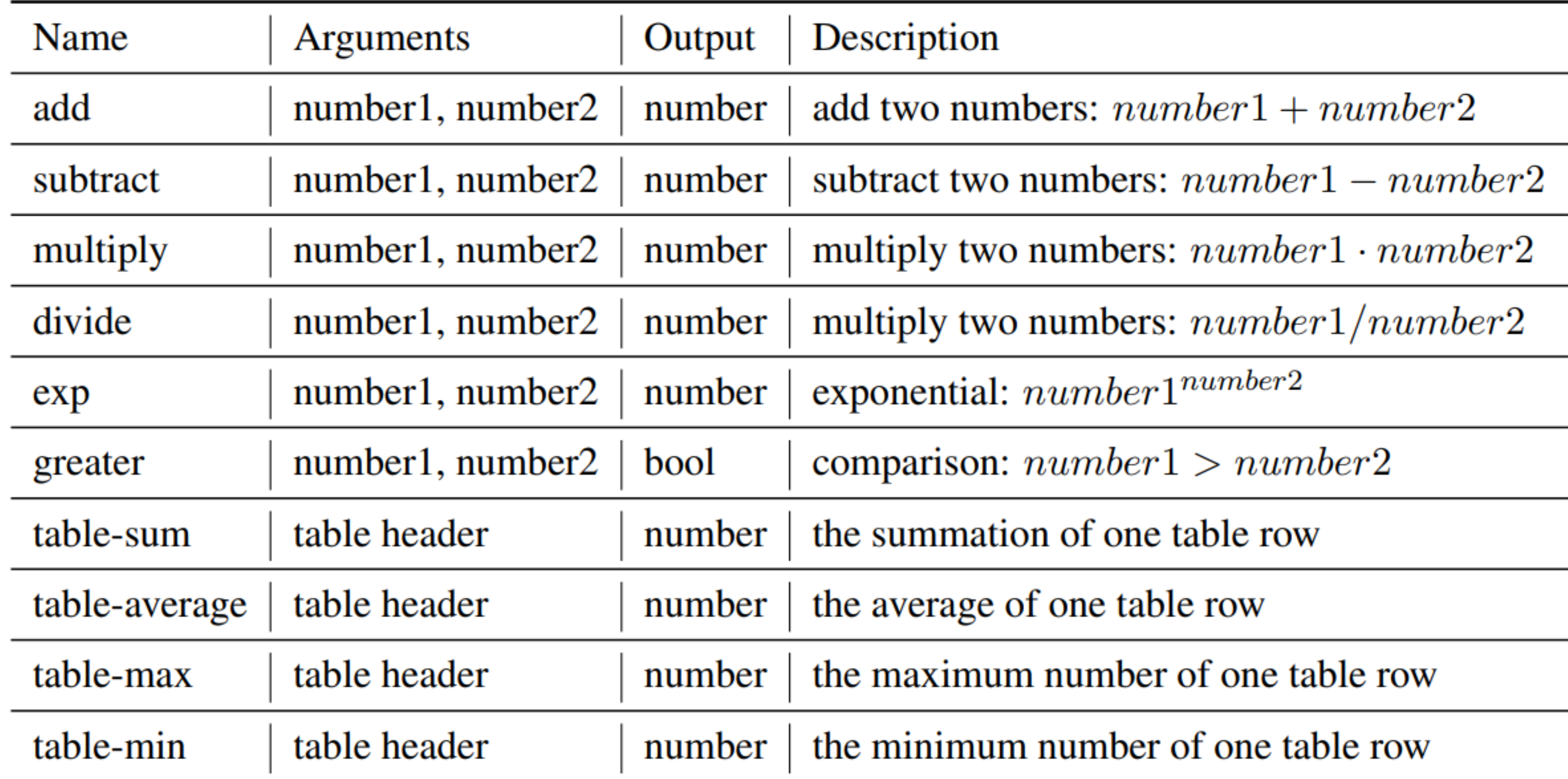}
    \caption{The definitions of operations according to FinQA original paper \cite{chen2021finqa}}
    \label{fig:my_label}
\end{figure}


Different from the traditional evaluation indicator of Q\&A task, which only use accuracy for evaluation, financial Q\&A task needs additional higher requirements of interpretability and
transparency. Thus in this task, the author proposed the gold programs for evaluation, which replace all the arguments with symbols, and then evaluate whether the two symbolic programs are mathematically equivalent. For instance, the follwing 2 programs are been judged as equivalent:
$$add(k_{4},k_{3}), add(k_{1},k_{2}), subtract(\#1, \#0)$$
$$add(k_{1},k_{2}), add(k_{3},k_{4}), subtract(\#0, \#1)$$
Where $\#n$ denotes the result from the $n^{th}$ step.

An example of the FinQA data set is shown in Figure 2. According to the description of the financial document on the left, input the question \textit{"what is the net change in net revenue during 2015 for entergy corporation?"}, the task needs to get the final answer and the corresponding resoning program. 
\begin{figure*}[htbp]
    \centering
    \includegraphics[width=\textwidth]{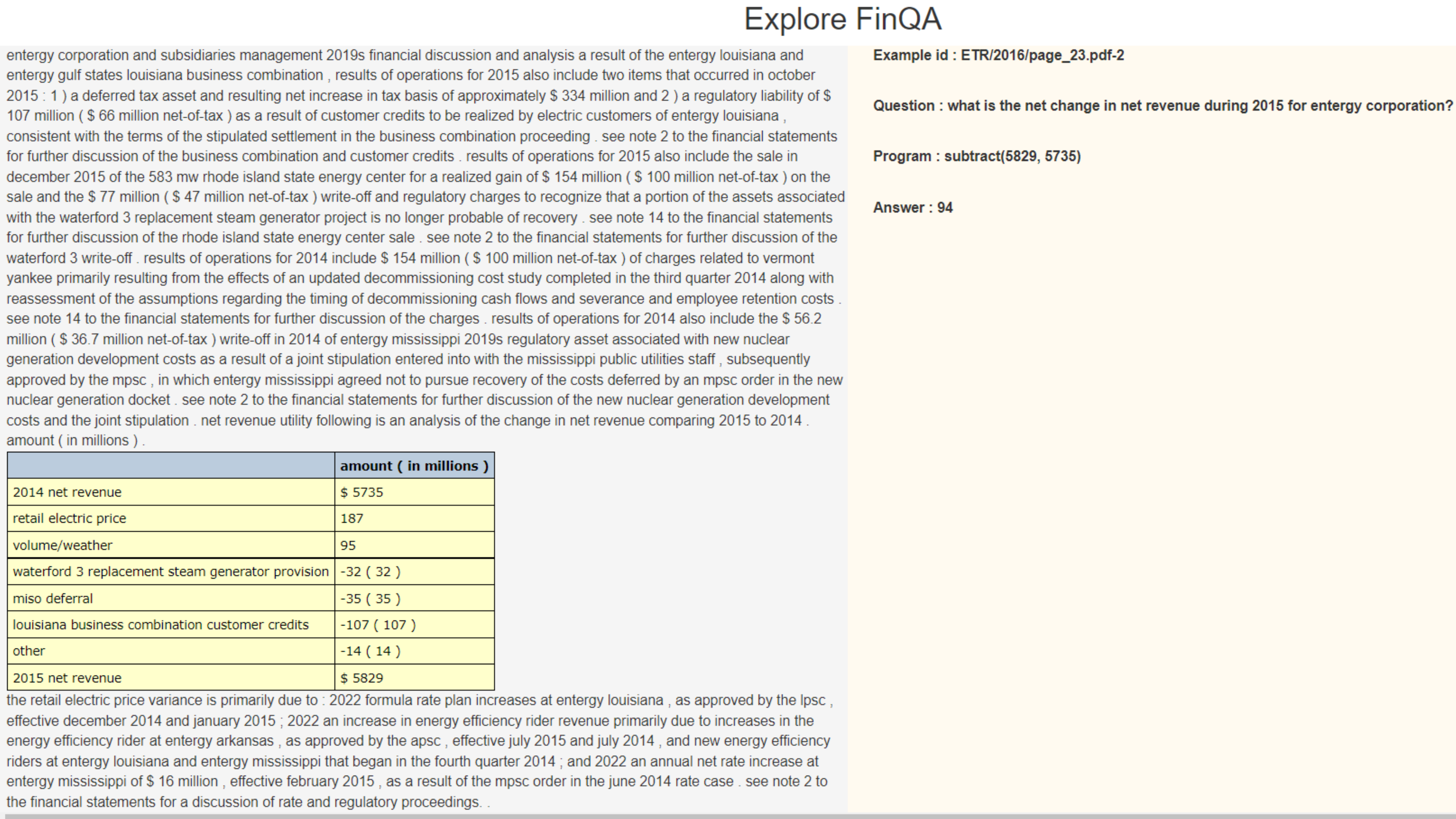}
    \caption{An example of the FinQA data set}
    \label{fig:my_label}
\end{figure*}
\subsection{The FinQANet Framework}
In the FinQA dataset, the author proposes the FinQANet framework, which uses the retriever to retrieve the supporting facts of financial documents, returns the sorted top $K$ supporting facts, and then uses the generator to generate the calculation formula to get the final answer $A$.

\subsubsection{Retriever}
Given the overall long length of the input financial documents, existing Q\&A models are often difficult to handle. Therefore, the author proposes to first extract some supporting facts from the input file, and then concatenate these supporting facts with the problem, and train a classifier based on the pretrained language model BERT \cite{devlin2018bert}. Then the author proposes to reorder these supporting facts and take the top $n$ supporting facts to enter the second stage of the retriever. In the second stage, the authors use a fixed-size sliding window for browsing and tagging, and finally as input to the generator.

\subsubsection{Generator}
In the generator part, the main goal is to use the supporting facts extracted by the retriever to generate the calculation formula and finally get the answer. The author divides the input into question tokens ${e_i}$, first encodes ${e_i}$ with the pretrained language model to get ${h^{e}_{i}}$, as well as the embeddings of special token ${h^{e}_{s}}$ from DSL and the step memory token ${h^{e}_{m}}$ (to denote the results from previous steps, like \#1). Finally all the token embeddings are denoted as $H=[h^{e}_{i}; h^{e}_{s},h^{e}_{m}] $. And then uses LSTM (Long Short Term Memory) for decoding. During inference, the authors use masks at each decoding step to guarantee the correct structure of the generated program. In the retrieval phase, the authors use the retriever results from the training set and superimpose the gold facts as the overall input.

\section{Methodology}
This submission mainly uses the pre-trained language model DeBERTa (Decoding-enhanced BERT with Disentangled Attention), and makes some fine-tuning on its basis. 

The pre-trained model was first used in the field of computer vision. It originated from AlexNet \cite{krizhevsky2012imagenet}, which won the first place in the Imagenet image recognition competition in 2012. Subsequently, BERT \cite{devlin2018bert} proposed by Google in the second half of 2018 has the significance of creating an era in the field of natural language processing. One of the advantages of the pre-trained model is that it can be used to build an effective model on a large amount of data of other similar tasks and then transfer to the target task, thus solving the problem of insufficient training data for the target task.

DeBERTa made the following improvements on the basis of BERT:

\subsection{Disentangled attention}
In BERT, each token in the input layer is represented by a vector, which is a simple combination of the token embedding, segment embedding and position embedding. However, in DeBERTa, each token is represented by two vectors, which encode the content and position respectively. According to the content and relative position of the token, the dispersion matrix is used to calculate the attention weight between tokens, which means the attention weights of token-pairs depend not only on their content, but also on their relative positions. 

In the process of calculating the attention weight, assuming that the word at position $i$ uses two vectors $h_i$ and $p_{i|j}$ to represent its content and the relative position of the token at position $j$, then the cross-attention $A_{i,j}$ between token $i$ and $j$ can be calculated as:
$$A_{i,j} = {h_{i}, p_{i|j}}\times{h_{j},p_{j|i}}^{T}$$
$$= h_{i}h_{j}^{T}+h_{i}p_{j|i}^{T}+p_{i|j}h_{j}^{T}+p_{i|j}p_{j|i}^{T}$$

In which needs to use the similarity between content and content (independent of position), the similarity between content and position, and the similarity between position and position.

Taking the single-head attention as an instance, the original self attention can be calculated through:
$$Q = hW_{q}, K = hW_{k}, V = hW_{v}, A = \frac{QK^{T}}{\sqrt{d}}$$
$$h_{o} = softmax(A)V$$

Where $h$ represents the input hidden victors, $h_{o}$ represents the output of self attention. $W_{q}$, $W_{k}$, $W_{v}$ are the  projection matrices, while $A$ is the attention matrix, and $d$ the dimension of hidden states.

On the basis of the former, the disentangled self-attention with relative position bias is shown as follows:
$$Q_{c} = hW_{q,c}, K_{c} = hW_{k,c}, V_{c} = hW_{v,c}$$
$$Q_{r} = pW_{q,r}, K_{r} = pW_{k,r}$$

In which $Q_{c}$, $K_{c}$ and $V_{c}$ represent the content vectors generated through projection matrices $W_{q,c}$, $W_{k,c}$ and $W_{v,c}$, and $Q_{r}$ and $K_{r}$ represent the relative position embedding
vectors generated through projection matrices $W_{q,r}$ and $W_{k,r}$. Then the attention matrix $\tilde{A}_{i,j}$ represented the attention score from token $i$ to token
$j$ is:
$$\tilde{A}_{i,j} = {Q_{i}^{c}K_{j}^{c}}^{T}+{Q_{i}^{c}K_{\sigma(i,j)}^{r}}^{T}+{K_{j}^{c}Q_{\sigma(j,i)}^{r}}^{T} $$
$$h^{o}=softmax(\frac{\tilde{A}}{\sqrt{3d}})V_{c}$$

Where $Q_{i}^{c}$ represent the $i^{th}$ row of $Q_{c}$, and $K_{j}^{c}$,$K_{\sigma(i,j)}$, $Q_{\sigma(j,i)}^{r}$,are in the same manner.

\subsection{Enhanced mask decoder }
Similar to BERT, DeBERTa is also pre-trained using MLM, a fill-in-the-blank task where a model is taught to use a mask to mask the surrounding words and then predict what the masked token should be. BERT uses the content and location information of context words for MLM. The attention mechanism already considers the content and relative positions of contextual words, but not the absolute positions of these words, which are critical for prediction in many cases. Therefore, DeBERTa has an enhanced mask decoder, which introduces absolute word position embeddings before the softmax layer, where the model decodes hidden words based on the aggregated contextual embeddings of word content and position. The comparation between BERT decoder and enhanced mask decoder of DeBERTa is shown in Figure 3.

\begin{figure}
    \centering
    \includegraphics[width=7.5cm]{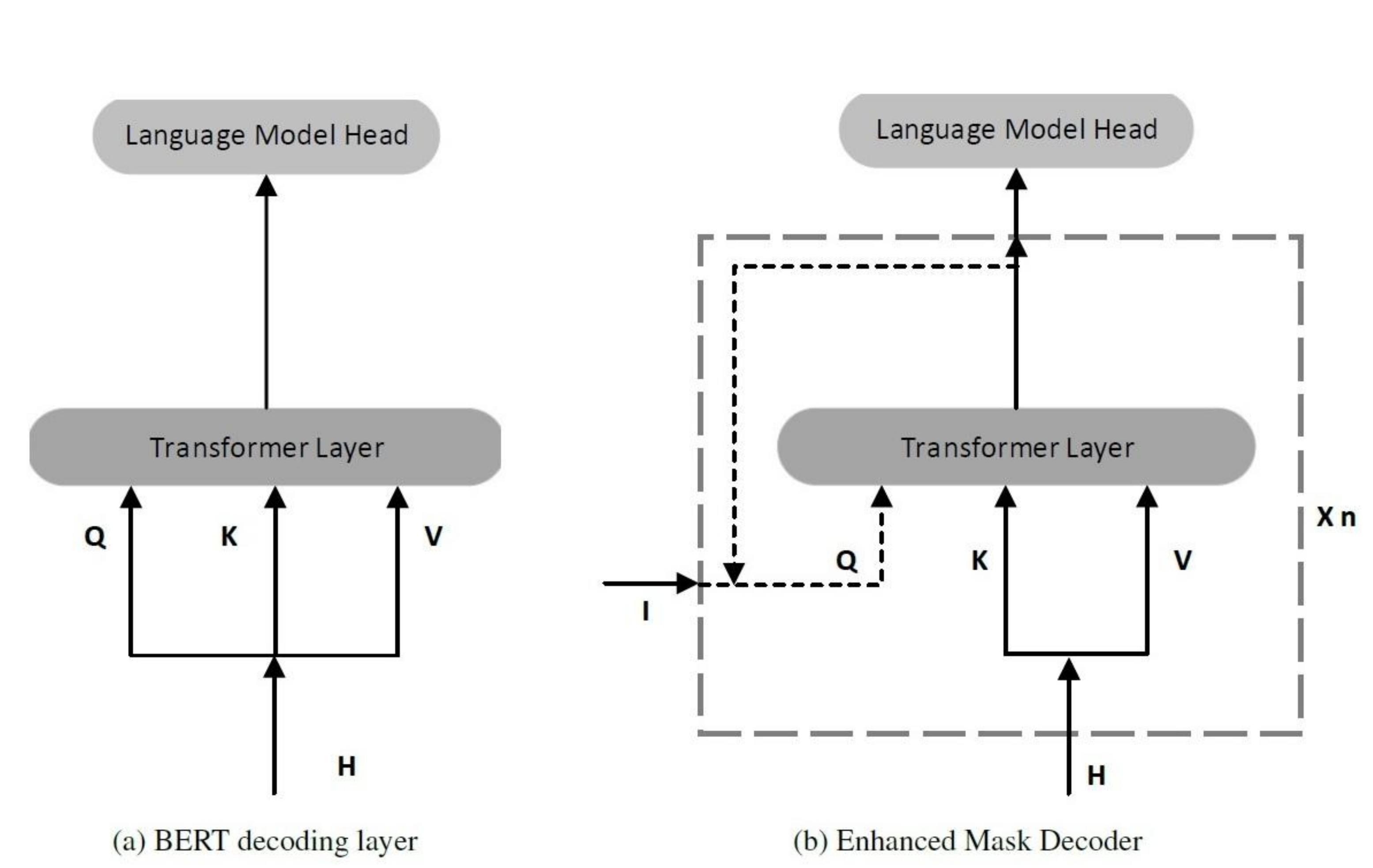}
    \caption{The comparation between BERT decoder and enhanced mask decoder of DeBERTa \cite{he2020deberta}}
    \label{fig:my_label}
\end{figure}
The left side of Figure 3 is the architecture of the transformer layer of BERT, which constructs $Q$, $K$ and $V$ from $H$, and then passes to the next transformer layer. The DeBERTa enhanced mask decoder on the right side of Figure 3 adds absolute position information $I$. For each layer of transformer layer, DeBERTa will take the absolute position information as input and participate in the calculation of the attention score.

\subsection{Virtual adversarial training}
DeBERTa proposed to use Scale-invariant-Fine-Tuning (SiFT) for virtual adversarial training. SIFT is a trick used to improve the training effect during fine tuning. It can be regarded as a regularization method similar to regularizer and dropout. When fine-tuning DeBERTa to downstream NLP tasks, SiFT first vector-normalizes the word embeddings, and then perturbs the normalized embedding vectors. This operation can greatly improve the performance of the model, especially for larger DeBERTa models.

\subsection{Our tricks}
our team has made fine-tuning and some additional tricks based on DeBERTa (Decoding-enhanced BERT with Disentangled Attention). The specific steps are as follows:
 Firstly we used MLM to do a pre-training on financial data, the backbone model we used is \textit{\textbf{DeBERTa v3 large}}\footnote{https://huggingface.co/microsoft/deberta-large}.
 
Secondly, We made some improvements to the Retriever in FinQA Framework. We have added multi-model fusion in the Retriever. Multi-model fusion is an optimization method that breaks through the bottleneck of model generalization and obtains the optimal solution by synthesizing the advantages of each model. In order to cooperate with the multi-model fusion on Retriever, we choose to combine the training data for enhancement. During the process of combination, we tried \textit{train}\footnote{train, test, dev represent the train.json, test.json and dev.json file in dataset respectively} only, \textit{train+test} and \textit{train+dev} respectively, and found that the model trained by \textit{train+dev} is prone to overfitting. Thus we choose to use train and train+test as training data, and create differences by changing the negative sample ratio $neg\_rate$.

According to the results of our simple preliminary experiment, the better the result predicted by the Retriever, the better the result obtained by the Generator. Therefore, we simply average the prediction results of the model weights obtained by training in the following three combinations, and provide them to the Generator for training and prediction, so as to improve the results of the Generator:
\begin{itemize}
    \item train+test,neg\_rate=3
    \item train,neg\_rate=3
    \item train,neg\_rate=4
\end{itemize}

Similarly, on the Generator, we also fused 12 models. The combinations we use are \textit{train}, \textit{train+test} and that with different seeds. Then the training data combination of the Generator and the training data combination of the Retriever are matched in pairs to form a differentiated model series, which lays the foundation for the subsequent integration. In the fusion stage of the generator, we tried simple average fusion, but the effect was not satisfactory, so we finally set different weights for each model to fuse.

Our team achieved an execution accuracy of 68.99 and a program accuracy of 64.53, ranking first in the FinQA Challenge\footnote{https://finqasite.github.io/challenge.html}.

\section{Conclusion}
As a rising star in the field of natural language processing, question answering systems (Q\&A Systems) are widely used in all walks of life. Compared with other scenarios, the application in financial scenario has strong traceability and interpretability requirements. In addition, since the demand for artificial intelligence technology has gradually shifted from the initial computational intelligence to cognitive intelligence. This research mainly focuses on the financial numerical reasoning dataset - FinQA. In the shared task, the objective is to generate the reasoning program and the final answer according to the given financial report containing text and tables. We use the method based on DeBERTa pre-trained language model, with additional optimization methodsincluding multi-model fusion, training set combination on this basis. We finally obtain an execution accuracy of 68.99 and a program accuracy of 64.53, ranking No. 4 in the FinQA Challenge.

\end{document}